\documentclass[letterpaper,10pt,conference]{ieeeconf}
\IEEEoverridecommandlockouts
\usepackage[T1]{fontenc}
\usepackage[utf8]{inputenc}
\usepackage{amsmath,amssymb}
\usepackage{booktabs,array,tabularx}
\usepackage{graphicx}
\usepackage{url}
\newcolumntype{Y}{>{\raggedright\arraybackslash}X}
\title{\LARGE\bf Compact Visuotactile World Models for Lifting:\protect\\
Prediction, Reward Alignment, and Force Constraints}
\author{Qinzhen Ma%
\thanks{Qinzhen Ma is with Rice University, USA. {\tt\small qm18@rice.edu}}}
\begin{document}
\maketitle
\thispagestyle{empty}
\pagestyle{empty}

\begin{abstract}
Accurate tactile forecasts need not improve force-constrained control.
We study a 652,157-parameter action-conditioned visuotactile world model with
matched behavior cloning, policy learning in imagination, independent
reactive implicit Q-learning, and model-assisted force feedback.
A fixed protocol executes 34 policies on 120 fresh MuJoCo environments
spanning geometry and physical-parameter shifts, plus 324 independently
replayed action branches on 12 additional ID environments. Visuotactile dynamics reduce force action-effect MAE from 0.413 N for persistence to 0.338 N. Model-assisted feedback raises ID force-budgeted success from 73.3\% to 93.3\%, with paired difference \mbox{$+20.0\,[+6.7,+33.4]$} percentage points (95\% CI), with the difference occurring during scripted lowering. Its pooled difference is \mbox{$+3.9\,[-4.5,+11.7]$} points. Imagined RL achieves 11.9\% pooled joint success versus 25.0\% for reactive IQL. 
An empirical tactile-residual stress test adds 330 executions.
The evidence concerns rigid-box lifting after a common approach, without
physical-robot transfer or a closed-loop safety guarantee.
\end{abstract}

\section{Introduction}
Vision describes object geometry and motion, whereas touch provides local
contact information. A compact model combining these signals could guide
lifting under a force budget. Yet favorable prediction metrics can be
misleading: contact forces may vary slowly enough for persistence to work,
small action effects can be obscured by state-estimation bias, and a policy
can complete a lift while exceeding its force limit. A task reward that
saturates before the required height adds a separate source of failure.

We ask which parts of a visuotactile control pipeline produce measurable
decision value. Does adding touch help when task reward is fixed? Does
optimizing a policy in a learned model improve its exact behavior-cloned
initialization? Does model-based candidate selection add value to a shared
force-feedback controller, or is the measured-force guard responsible for the
apparent improvement? These questions require both paired executed outcomes
and direct tests of action-dependent predictions.

Our study follows an exploratory lifting pilot in which changing a native
reward to a height-aligned reward improved lifting but did not establish
superior force-budgeted performance. The new experiment freezes data splits,
controller gains, test conditions, and analysis before reading its test
outcomes. It expands the geometry and physical-parameter supports, compares
independent reactive learning methods, and explicitly separates representation
learning, imagined policy improvement, and feedback protection. It also
tests forecasts on alternative actions actually executed from identical
simulator prefixes, rather than relying only on recorded-action error.

The contribution is an empirical diagnosis in a compact-model regime,
not a new modality-fusion architecture or safety theorem. First, the matched
modality and initialization comparisons identify what the tested sensing and
policy-improvement stages contribute. Second, a common feedback base with
model/guard ablations measures the selector's incremental effect. Third,
independently replayed local branches distinguish absolute forecast accuracy
from action-effect prediction and constrained choice. The model-assisted selector improves ID and held-out-geometry outcomes, while physical-parameter shifts remove its demonstrated advantage. Better local action effects coexist with inconclusive joint-success gains from imagined policy optimization; WM-RL VT has lower pooled success than reactive IQL VT.
All failed approaches and force peaks during setup and lowering remain in
the primary evaluation. Public tactile measurements supply an additional
error-perturbation diagnostic, with its cross-sensor and timing assumptions
stated separately.

\begin{figure*}[t]
\vspace*{5pt}
\centering
\includegraphics[width=\textwidth]{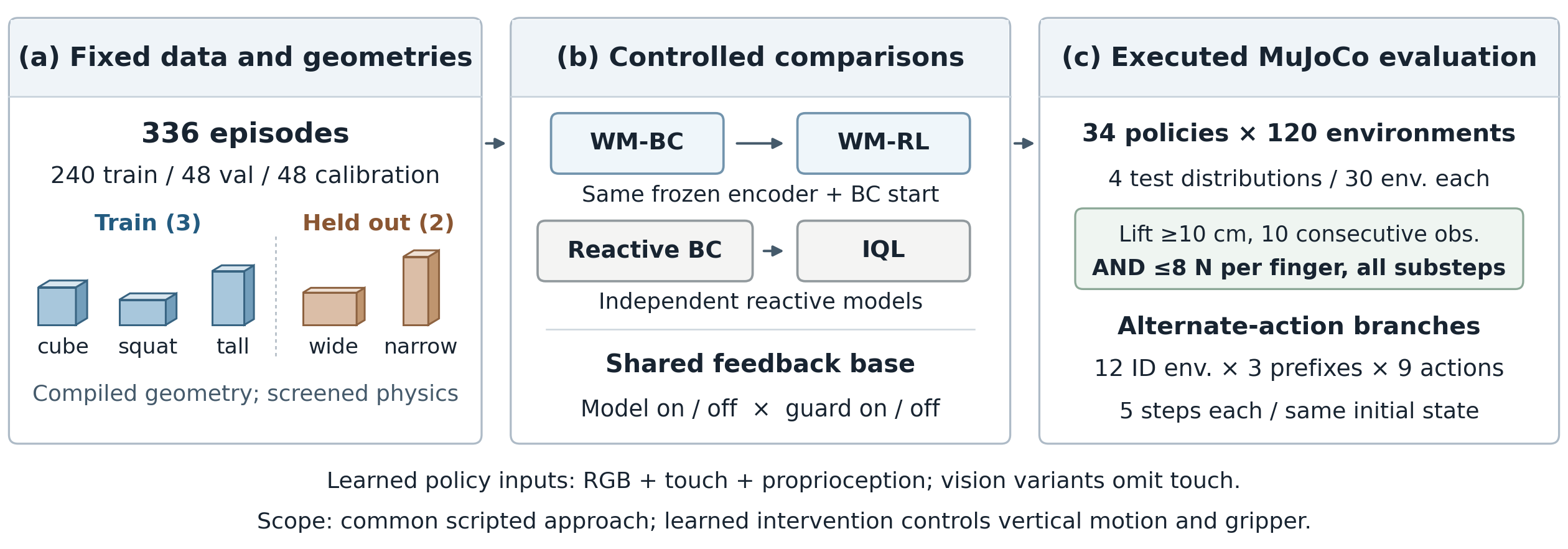}
\caption{Controlled experimental comparisons. Training and test environments
are disjoint. Controllers share an approach and a two-dimensional intervention
space. Local action branches are independently reset and replayed in MuJoCo;
their observed outcomes are never replaced by model predictions.}
\label{fig:protocol}
\end{figure*}

\section{Related Work}
Tactile predictive control predates current world-model terminology:
Tian \emph{et al.} use learned tactile prediction for manipulation~\cite{tactilempc}.
VT-WM combines visuotactile prediction and planning~\cite{vtwm}; FeelWorld
studies hierarchical contact prediction~\cite{feelworld}; and OmniVTA combines
predictive tactile features with reactive feedback and ablates their
contributions~\cite{omnivta}. We therefore do not claim that combining vision,
touch, forecasting, or a reflex is itself new. Our focus is the measured
relationship between compact force forecasts, executed alternative actions,
and a whole-episode force-budgeted outcome.

TD-MPC2 supplies latent-model building blocks and a complete model-based
learning system~\cite{tdmpc2}, while Dreamer learns policies in imagined
trajectories~\cite{dreamer}. We adapt selected open-source TD-MPC2 convolutional,
MLP, and SimNorm components; our frozen-model actor--critic is not a
reproduction of either complete algorithm. Implicit Q-learning (IQL) offers
an offline reactive alternative using observed transitions~\cite{iql}.
Our task-specific pixel IQL implements its expectile-value and
advantage-weighted policy objectives, with independent encoders. The
pipeline comparison is informative but does not hold every architecture and
optimization choice constant.

Prediction and decision quality are conceptually distinct in model-based
reinforcement learning~\cite{valueequivalence}. Split conformal calibration
can account for trajectory-level errors under appropriate exchangeability
conditions~\cite{conformal}; it does not automatically cover adaptive
closed-loop actions. Sparsh/TacBench supplies public tactile images paired
with force measurements~\cite{sparsh}. We use those records for a separate
sensing analysis and a transparent error perturbation, without treating them
as real-robot executions of our policy.

\section{Task and Experimental Protocol}
\subsection{Observation, action, and measurement}
The environment extends robosuite Panda Lift~\cite{robosuite} in
MuJoCo~\cite{mujoco}. Each episode contains 150 actions at 20 Hz; each action
is integrated through 25 physics substeps of 0.002 s. The robot uses a
continuous gripper-rate interface. Three $64\times64$ external RGB frames,
three six-dimensional tactile readings, and three nine-dimensional
proprioceptive readings form the policy observation. Each finger supplies
one normal and two signed tangential force aggregates. Proprioception
contains end-effector pose and finger positions. These are rigid-simulator
contact-force signals, not rendered optical tactile images.

Every controller receives the same privileged scripted approach at actions
0--43, using object geometry. Policies warm their observation history at
actions 42--43 and control only vertical displacement and gripper rate at
actions 44--134. Horizontal and rotational increments are zero in this
interval for every method. Actions 135--149 perform common scripted lowering.
The policy interface contains no object height, pose, mass, friction,
success label, or simulator reward. The adaptive controller additionally
receives the common intervention-start indicator, not a task-success signal.
This protocol studies grasp closure and lifting, not unrestricted reaching
or six-dimensional manipulation.

The strict lift outcome requires object height relative to reset to reach
0.10 m at ten consecutive observations before lowering. The first and last
of these samples are 0.45 s apart; height is not checked continuously between
them. Let $L_i$ denote this event and $p_{itj}$ the normal-force maximum for
finger $j$ over the substeps of action $t$. Our primary outcome is
\begin{equation}
J_i=\mathbf{1}\{L_i\}\,\mathbf{1}\{\max_{t,j}p_{itj}\leq8\ \mathrm{N}\}.
\label{eq:joint}
\end{equation}
The force maximum includes the complete episode, including approach and
lowering. We report $L_i$, $J_i$, and force violations separately; all failed
approaches remain in their denominators.

\subsection{Geometries, distributions, and data}
Objects are compiled as rigid boxes before simulation. Three training
geometries have half dimensions, in meters, of a fixed near-cube cuboid
$(.021184,.021703,.020551)$ (rounded; compile seed 20270915), a squat box
$(.026,.020,.014)$, and a tall box $(.018,.018,.030)$.
Two held-out geometries are a wide box
$(.030,.015,.018)$ and a narrow tall box $(.014,.020,.038)$.
This tests aspect-ratio and size generalization within boxes, not new object
categories or deformable materials.

ID masses are $\{.06,.12,.25\}$ kg and friction coefficients are
$\{.60,.75,.90\}$. Physical-parameter OOD uses masses
$\{.30,.40,.50\}$ kg and friction $\{.30,.40,.50\}$.
Actuator caps are $\{6,12,20\}$ N in all domains. A static antipodal
grasp screen requires
\begin{equation}
\frac{2\mu\min(8,c)}{mg}\geq1.5,
\label{eq:screen}
\end{equation}
with $g=9.81$ m/s$^2$ and a conservative aperture check
$2\sqrt{h_x^2+h_y^2}+2\|\delta_{xy}\|+.002\leq.079$ m,
where $h_x,h_y$ are box half dimensions and $\|\delta_{xy}\|\leq.0035$ m
is the grasp offset. Here $c$ is a nominal actuator-force proxy, not an
exact contact-force clamp. We sample the cap uniformly, then a feasible
mass/friction pair uniformly conditional on that cap. This conservative
screen uses a simplified static grasp model and cannot guarantee dynamic
feasibility or controller success.

We collect 240 training, 48 validation, and 48 calibration episodes, balanced
across three geometries and a 50:50 mix of script and 5 N force feedback.
The collection enforces the same two action dimensions during intervention
as the evaluation. This avoids fitting a two-action Q function to transitions
actually produced by unmodeled horizontal/rotational action components.
Normalization uses training episodes alone. All trained methods use the same
recorded collection, with seeds 0, 1, and 2 and no online retraining.

There are 30 new environments in each of ID, geometry OOD, physics OOD, and
combined OOD. Every policy executes separately after resetting the same
environment seed. The four domains use separate seed ranges: pairing is
within each environment across methods, not a matched intervention on
physical parameters across domains. A further 12 disjoint environments are
reserved for branch diagnostics. The protocol is locally prospective, not
publicly preregistered. Before test execution, a documented causal review
added six predictive-encoder BC controls and vision-model branch forecasts
to the original 28-configuration plan. No test outcomes selected weights,
gains, reward coefficients, or included environments.

\section{Models and Controlled Comparisons}
\subsection{Compact world model}
A visual convolutional encoder produces 256 features; tactile and
proprioceptive MLPs produce 64 and 32 features. A fused 128-dimensional
SimNorm latent is propagated by a transition MLP with two 384-unit hidden layers:
\begin{equation}
z_t=e_\theta(o_{t-2:t}),\qquad
\widehat z_{t+k+1}=d_\theta(\widehat z_{t+k},a_{t+k}).
\end{equation}
Heads predict six endpoint-force components, two substep-peak normal
forces, relative object height, support, finger contact, native reward,
and an auxiliary $16\times16$ RGB target. Object state and substep peaks
are training targets only. The model has 652,157 parameters and starts
from random weights. Vision and visuotactile variants have identical
architectures and parameter counts; vision masks normalized touch to zero.

Training supervises the initial decoded state and five future states.
Normalized Huber losses for endpoint force $F$, peak force $p$, height $h$,
and reward $r$ combine with binary support/contact losses, RGB error, and
latent consistency to a stop-gradient EMA encoder:
\begin{equation}
\begin{split}
\mathcal L={}&\mathcal L_F+\mathcal L_p+\mathcal L_h
+0.25(\mathcal L_S+\mathcal L_C)\\
&+0.5\mathcal L_r+0.5\mathcal L_I+20\mathcal L_z.
\end{split}
\end{equation}
All six models train for 25 epochs with batch size 64, AdamW learning rate
$3\times10^{-4}$, weight decay $10^{-4}$, gradient clip 20, and EMA rate
0.01. Training/validation window strides are 3/5. Checkpoint selection uses
validation loss excluding the EMA latent term, avoiding selection by
target-encoder lag. No pretrained visual or tactile weights are used.

\subsection{Predictive BC and imagined policy improvement}
We first fit a 33,282-parameter, two-hidden-layer actor to actions from
encoded training histories while freezing the world model. The saved
actor is \emph{WM-BC}. Its exact weights initialize \emph{WM-RL}.
Actions are bounded by training-action ranges. The actor learns five-step
imagined returns with height-aligned reward
\begin{equation}
\begin{split}
\widetilde r_k={}&\mathrm{clip}(\widehat h_k/0.10,0,1)\\
&-\frac12\sum_{j=1}^{2}\left(\frac{[\max(0,\widehat p_{kj})-8]_+}{8}\right)^2
-0.005\,\overline{u_k^2}.
\end{split}
\label{eq:reward}
\end{equation}
The actor minimizes negative discounted return plus twice the mean squared
deviation from a frozen BC prior. A value network regresses detached
imagined returns; an EMA target critic provides a terminal bootstrap,
with gradient through the terminal latent and zero bootstrap at the
intervention boundary. Discount is 0.95. Each run uses 1,000 BC and 1,500
imagined updates, batch 256, actor/value learning rates
$2\times10^{-4}/3\times10^{-4}$, and target-value rate 0.01.
Encoded starts span actions 44--130 at stride two. Simulator state and
reward are never queried during imagined updates.

The paired WM-RL minus WM-BC contrast measures the subsequent policy
improvement procedure, including its extra 1,500 updates, under the same
frozen predictive representation. It does not isolate representation
learning or match optimization budgets; WM-BC receives no extra updates.

\subsection{Independent reactive BC and IQL}
Reactive policies receive the same observation history and action dimensions
but train fresh encoders without a transition model, latent rollouts, or
predicted rewards. Actor, twin-Q critic, and value each use a separate
encoder. BC fits a Gaussian action likelihood and deploys the bounded mean.
IQL fits an expectile value to the minimum target Q, temporal-difference
critics using observed next-state reward, and an advantage-weighted action
likelihood~\cite{iql}. Its reward is Eq.~\eqref{eq:reward} evaluated with
actual next-state height and interval peaks in the training records.

Each modality and seed uses 2,000 BC updates followed by 10,000 IQL updates,
batch 256, Adam learning rate $3\times10^{-4}$, expectile 0.7, inverse
temperature 3, maximum advantage weight 100, discount 0.95, target EMA
0.005, and gradient clip 10. BC selects minimum validation normalized-action
MSE; IQL uses the fixed final update. Validation never becomes an online
reward or test-selection signal. The deployed reactive actor has 184,692
parameters; total trainable IQL parameters are 785,751, plus 350,578
nontrainable target-critic parameters. WM versus reactive learning is a
whole-pipeline comparison with different objectives, encoders, and compute.

\subsection{Separating model selection from measured-force protection}
The script baseline retains its scripted vertical/gripper commands.
The force-feedback baseline replaces gripper rate with
$\mathrm{clip}(0.12(5-\overline F_n),-.25,.35)$ while retaining the
scripted vertical command. A second, common reactive base uses this
gripper feedback and an end-effector-relative 0.14 m vertical ramp after
18 grasp steps, increasing over 50 steps. Vertical gain is 12 and command
magnitude is limited to 0.6. Upward motion stops when either measured
finger normal is below 0.25 N. This base uses proprioception and touch;
it never observes actual object height.

A measured-force guard opens at rate $-0.25$ and halts upward motion if
either current normal reaches 7.25 N. It cannot retroactively prevent a
substep peak and is not a hard constraint guarantee. The model-assisted
selector forms local vertical/gripper offsets
$\{0,-.06,.06\}\times\{0,-.10,.10\}$. Clipping, the base's contact/pre-lift
conditions, and the current optional guard leave up to nine unique actions.
Each is repeated for five open-loop predicted steps; future guard decisions
are not simulated.
A current-force offset corrects decoder bias:
\begin{equation}
\widehat p^{\mathrm{anc}}_{kj}=
\left[\widehat p_{kj}+F_{n,tj}-\widehat F_{n,tj}\right]_+.
\end{equation}
Among candidates with maximum anchored five-step peak at most 8 N, it
maximizes $\mathrm{clip}(\widehat h_5/.10,0,1)
-.02\,\overline{p^{\mathrm{anc}}}/8-.002\|a-a_{\mathrm{base}}\|^2$,
where the mean covers five steps and both fingers and $a_{\mathrm{base}}$
includes the optional guard. If none is feasible, it uses the base.
This offset is not conformal
calibration. The four ablations retain/remove the selector and guard
independently. Model-assisted modes use three visuotactile models;
reactive modes are deterministic single configurations. All gains and
candidate grids were fixed before the new outcomes were read.

\section{Results}
\subsection{Executed lifting and force-budget success}
The main study comprises 4,080 executions of 34 policy configurations on
120 independent environment conditions. For each method group, we first
average its three fixed training-seed policies within an environment, then
average environments; deterministic baselines use one policy.
Confidence intervals use 2,000 paired environment-bootstrap resamples
within each domain, seed 20401010, with the same draws across methods.
They condition on the fitted policies. Training-seed variation is reported
separately. Unadjusted intervals are descriptive, not simultaneous
certificates for multiple superiority claims.
Table~\ref{tab:control} reports all outcomes and Table~\ref{tab:contrasts} all ten fixed primary contrasts. Model-assisted feedback attains 93.3\% ID joint success versus 73.3\% for its shared reactive base: \mbox{$+20.0\,[+6.7,+33.4]$} percentage points (pp). The geometry-shift difference is \mbox{$+16.7\,[+3.3,+30.0]$} pp. Under physics/combined shifts, joint success is 55.6/53.3\%, versus 60.0/70.0\% for the reactive base. Pooled joint success is 68.9\% versus 65.0\%; its paired difference \mbox{$+3.9\,[-4.5,+11.7]$} pp includes zero. Pooled strict lifting falls from 97.5\% to 81.1\%. In ID and geometry OOD, the model-assisted and shared reactive controllers lift successfully in every execution, and every force violation occurs during common scripted lowering. The joint-success gains are expressed during lowering rather than as fewer active-phase force violations. Guard on/off variants have identical observed binary-outcome rates, despite 23 and 2 guard interventions across 360 model-assisted and 120 reactive executions, respectively. These sparse activations do not establish a protective benefit.

With fixed height reward, WM-RL VT exceeds vision in ID joint success by \mbox{$+7.8\,[+1.1,+14.4]$} pp, but the pooled difference \mbox{$+0.6\,[-2.5,+3.6]$} pp is inconclusive. Pooled IQL VT joint success is 25.0\%, compared with 11.9\% for WM-RL VT; WM-RL minus IQL is \mbox{$-13.1\,[-18.9,-7.8]$} pp. WM-RL VT increases strict lifting over its exact WM-BC initialization (62.2\% versus 29.4\%), yet its joint gain is only \mbox{$+1.4\,[-3.9,+6.4]$} pp. The corresponding vision and predictive-versus-reactive BC contrasts also include zero in the pooled joint metric.

Training-seed variation remains material: individual ID joint rates span 20.0--33.3\% for WM-RL VT and 23.3--50.0\% for IQL VT; all three model-assisted policies attain 93.3\% ID. The bootstrap intervals condition on these fitted policies rather than resampling training.
\begin{table*}[t]
\vspace*{5pt}
\centering
\caption{Executed outcomes (\%): joint force-budgeted success / strict lift / force violation (J/L/V). Every domain contains the same fixed 30 environments across methods. $n$ counts policies: learned groups average seeds 0, 1, 2 within each environment; single controllers have $n=1$. V/VT denote vision/visuotactile; FFB is force feedback. All setup and lowering failures and force peaks are included.}
\label{tab:control}
\begingroup
\fontsize{8}{9.2}\selectfont
\setlength{\tabcolsep}{2pt}
\renewcommand{\arraystretch}{1.10}
\begin{tabular*}{\textwidth}{@{\extracolsep{\fill}}lcrrrr@{}}
\toprule
Method & $n$ & ID & Geometry OOD & Physics OOD & Combined OOD \\
 & & J/L/V & J/L/V & J/L/V & J/L/V \\
\midrule
WM-BC V & 3 & 27.8/56.7/28.9 & 5.6/21.1/15.6 & 14.4/38.9/34.4 & 4.4/11.1/11.1 \\
WM-BC VT & 3 & 23.3/45.6/35.6 & 7.8/31.1/32.2 & 6.7/25.6/40.0 & 4.4/15.6/32.2 \\
WM-RL V & 3 & 20.0/76.7/75.6 & 23.3/54.4/50.0 & 1.1/80.0/97.8 & 1.1/47.8/76.7 \\
WM-RL VT & 3 & 27.8/83.3/57.8 & 18.9/70.0/66.7 & 1.1/50.0/85.6 & 0.0/45.6/93.3 \\
\addlinespace[2pt]
BC V & 3 & 15.6/56.7/57.8 & 5.6/11.1/27.8 & 2.2/13.3/63.3 & 1.1/8.9/42.2 \\
BC VT & 3 & 15.6/46.7/41.1 & 11.1/26.7/17.8 & 6.7/45.6/53.3 & 4.4/16.7/28.9 \\
IQL V & 3 & 27.8/72.2/54.4 & 15.6/42.2/34.4 & 15.6/36.7/60.0 & 5.6/31.1/51.1 \\
IQL VT & 3 & 35.6/82.2/56.7 & 33.3/72.2/41.1 & 17.8/61.1/63.3 & 13.3/61.1/52.2 \\
\addlinespace[2pt]
Script & 1 & 16.7/73.3/56.7 & 23.3/76.7/53.3 & 6.7/90.0/83.3 & 3.3/53.3/50.0 \\
FFB & 1 & 73.3/100.0/26.7 & 56.7/100.0/43.3 & 60.0/96.7/40.0 & 56.7/83.3/40.0 \\
\addlinespace[2pt]
Model+guard & 3 & 93.3/100.0/6.7 & 73.3/100.0/26.7 & 55.6/67.8/38.9 & 53.3/56.7/38.9 \\
Model & 3 & 93.3/100.0/6.7 & 73.3/100.0/26.7 & 55.6/67.8/38.9 & 53.3/56.7/38.9 \\
Reactive+guard & 1 & 73.3/100.0/26.7 & 56.7/100.0/43.3 & 60.0/96.7/40.0 & 70.0/93.3/26.7 \\
Reactive & 1 & 73.3/100.0/26.7 & 56.7/100.0/43.3 & 60.0/96.7/40.0 & 70.0/93.3/26.7 \\
\bottomrule
\end{tabular*}
\endgroup
\end{table*}

\begin{table}[t]
\vspace*{5pt}
\centering
\caption{All ten fixed joint-success contrasts: difference in percentage points and pointwise 95\% CI. ID: 30 environments; pooled: 120, stratified by domain. Intervals use 2,000 paired environment resamples, conditional on fitted policies. G denotes the measured-force guard; intervals are not multiplicity adjusted.}
\label{tab:contrasts}
\begingroup
\fontsize{8}{9.2}\selectfont
\setlength{\tabcolsep}{2pt}
\renewcommand{\arraystretch}{1.10}
\begin{tabular*}{\columnwidth}{@{\extracolsep{\fill}}lrr@{}}
\toprule
Contrast & ID & Pooled \\
\midrule
WM-RL VT $-$ V & \shortstack[r]{+7.8\\$[+1.1,+14.4]$} & \shortstack[r]{+0.6\\$[-2.5,+3.6]$} \\
IQL VT $-$ V & \shortstack[r]{+7.8\\$[-2.2,+17.8]$} & \shortstack[r]{+8.9\\$[+4.7,+13.3]$} \\
WM-RL VT $-$ IQL VT & \shortstack[r]{-7.8\\$[-22.2,+5.6]$} & \shortstack[r]{-13.1\\$[-18.9,-7.8]$} \\
Model+G $-$ Reactive+G & \shortstack[r]{+20.0\\$[+6.7,+33.4]$} & \shortstack[r]{+3.9\\$[-4.5,+11.7]$} \\
Model+G $-$ Model & \shortstack[r]{0.0\\$[0.0,0.0]$} & \shortstack[r]{0.0\\$[0.0,0.0]$} \\
Reactive+G $-$ Reactive & \shortstack[r]{0.0\\$[0.0,0.0]$} & \shortstack[r]{0.0\\$[0.0,0.0]$} \\
Model+G $-$ FFB & \shortstack[r]{+20.0\\$[+6.7,+33.4]$} & \shortstack[r]{+7.2\\$[-0.3,+15.0]$} \\
\addlinespace[2pt]
WM-RL V $-$ WM-BC V & \shortstack[r]{-7.8\\$[-20.0,+5.6]$} & \shortstack[r]{-1.7\\$[-6.4,+3.3]$} \\
WM-RL VT $-$ WM-BC VT & \shortstack[r]{+4.4\\$[-13.3,+20.0]$} & \shortstack[r]{+1.4\\$[-3.9,+6.4]$} \\
WM-BC VT $-$ BC VT & \shortstack[r]{+7.8\\$[-4.4,+22.2]$} & \shortstack[r]{+1.1\\$[-3.1,+5.6]$} \\
\bottomrule
\end{tabular*}
\endgroup
\end{table}

On the separate exploratory legacy cohort of ten ID environments, the added vision-height control gives 46.7\% strict lifting and 23.3\% joint success, versus 93.3\% and 33.3\% for VT with the same height reward. The joint difference is \mbox{$+10.0\,[-6.7,+33.3]$} pp. These original-distribution results are not pooled with the new study.

\subsection{Forecasts on a common recorded-action distribution}
Every model forecasts the same script-policy trajectories from the 120
test environments, using five steps, history three, and stride five.
Endpoint and interval-peak MAE average over horizons, windows, and
appropriate coordinates. VT diagnostics hold current tactile readings for
endpoint persistence and current decoded peak/height for their persistence;
the latter are not privileged online observations. Shuffled touch uses the
next episode's tactile history within the same test domain.

Fixed calibration radii are computed from 48 calibration episodes using
episode maxima over evaluated windows and future steps. Force uses
positive peak underprediction, maximized over fingers; height uses absolute
error. Each quantity receives error allowance $\alpha/2$ for nominal
joint coverage $1-\alpha=0.9$, with corrected split-conformal rank
$\lceil(n+1)(1-\alpha/2)\rceil$. Coverage here is empirical: calibration
mixes script and feedback policies whereas forecast tests use script,
and OOD groups introduce further shifts. Uncalibrated diagnostic
predictors do not inherit the world model's radius.
Table~\ref{tab:forecast} shows lower endpoint and peak errors for VT than vision in all four domains; their paired 95\% intervals exclude zero. ID endpoint/peak MAE decreases from 0.939/2.567 to 0.189/0.448 N. However, observed endpoint persistence achieves 0.106 N ID and lower endpoint error in every domain. All paired peak-error contrasts against decoded-peak persistence include zero. Shuffled touch increases both force errors in every domain. Height rollout does improve decoded-height persistence in ID, geometry, and physics OOD; the combined-shift interval includes zero. Thus the force-persistence finding does not extend to every predicted quantity.

VT nominal-90\% joint coverage is 90.0\% ID and 55.6--68.9\% across shifts. Its fixed upper force margins span 6.855--8.699 N across seeds, versus 14.679--15.479 N for vision, large relative to the 8 N task budget. Coverage and interval size must be considered together; neither supports a closed-loop safety claim.
\begin{table}[t]
\vspace*{5pt}
\centering
\caption{Common script-trajectory forecasts: 30 environments per domain, three fixed models. F/P: endpoint/interval-peak MAE (N). Persistence holds observed endpoint forces but current decoded peaks, not measured interval peaks. VT joint coverage (\%) uses fixed 48-episode calibration; coverage is empirical under policy/domain shift.}
\label{tab:forecast}
\begingroup
\fontsize{8}{9.2}\selectfont
\setlength{\tabcolsep}{2pt}
\renewcommand{\arraystretch}{1.10}
\begin{tabular*}{\columnwidth}{@{\extracolsep{\fill}}lrrrr@{}}
\toprule
Domain & Vision & VT & Persistence & VT joint \\
 & F/P & F/P & F/P & (\%) \\
\midrule
ID & 0.939/2.567 & 0.189/0.448 & 0.106/0.444 & 90.0 \\
Geometry & 1.159/2.508 & 0.631/1.134 & 0.293/1.195 & 55.6 \\
Physics & 1.230/2.977 & 0.402/0.789 & 0.133/0.788 & 68.9 \\
Combined & 0.772/1.690 & 0.361/0.668 & 0.132/0.704 & 65.6 \\
\bottomrule
\end{tabular*}
\endgroup
\end{table}

\subsection{Executed alternative-action diagnostic}
For each of 12 disjoint ID environments (four per training geometry), we
independently reset and replay
44, 78, or 112 common script/force-feedback actions, then execute nine
constant five-step branches around the feedback action. A branch begins
at that zero-based action index. Prefix RGB must match exactly; numerical
observation and simulator-state differences must be at most $10^{-7}$.
No partial snapshot is restored. Clipping duplicates are retained and counted.
The same 324 branches evaluate vision/visuotactile models with three
seeds, raw and force-anchored predictions, and persistence.
Raw vision excludes touch; anchored vision additionally receives measured
current normals for the same offset correction.
The primary force-persistence diagnostic holds the current measured normal
force; a previous-interval-peak variant is retained separately. Diagnostic
height persistence holds the true prebranch object height relative to reset,
an intentionally favorable reference unavailable to deployed policies.
Both assign zero action effects; privileged height supplies no action sensitivity.

\begin{figure}[t]
\vspace*{5pt}
\centering
\includegraphics[width=\columnwidth]{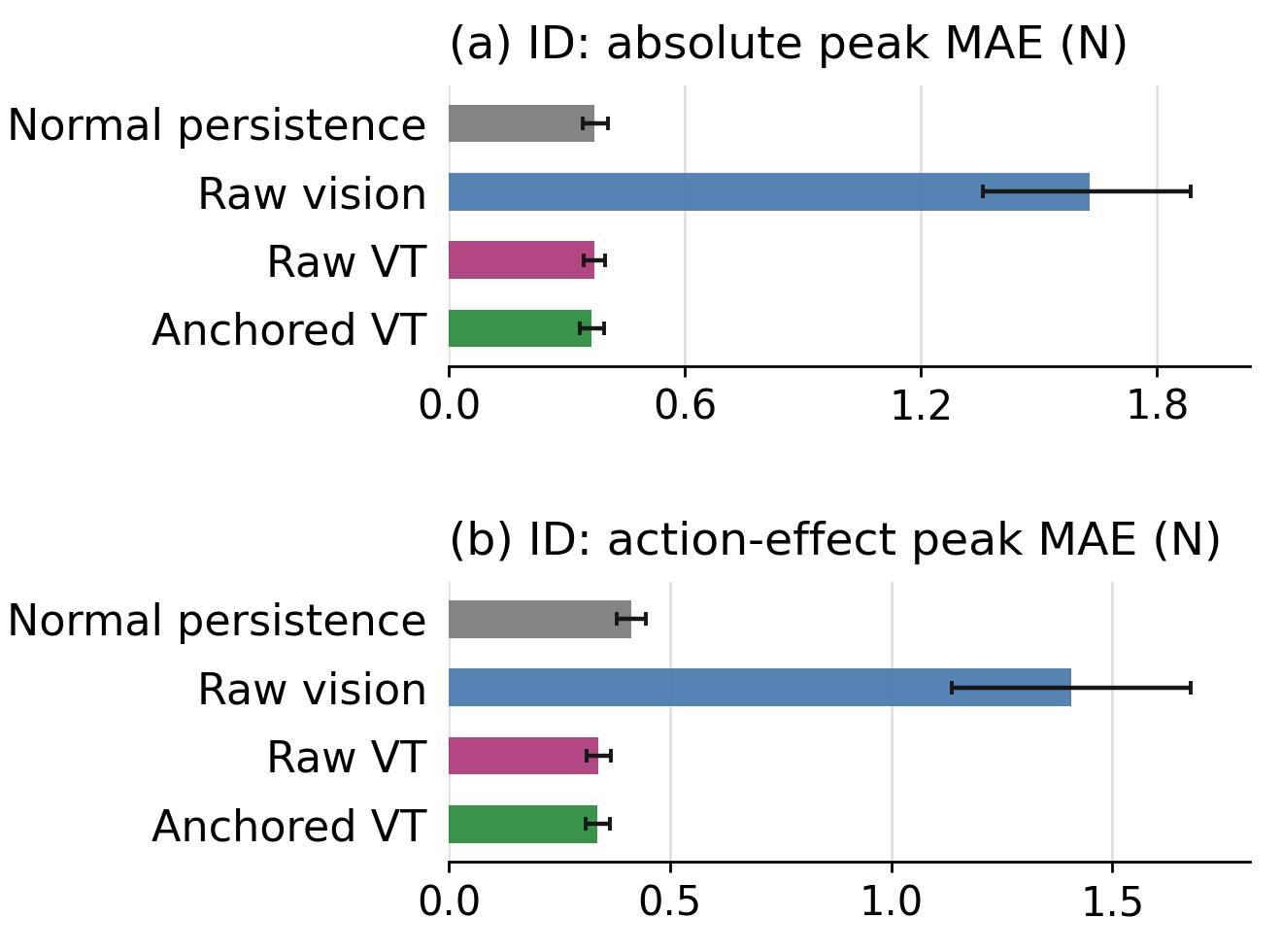}
\caption{Absolute force error and candidate-minus-center action-effect
error on independently executed ID branches. Model bars average three fixed
training seeds; intervals are 95\% CIs for each predictor's mean from 12
environment resamples, retaining all repeated branches.
Anchoring corrects current-force bias and is distinct from learning better
action effects.}
\label{fig:effects}
\end{figure}

For candidate $a$ and central action $a_0$, the force action-effect error
compares predicted and executed differences:
\begin{equation}
E_\Delta=\left|
(\widehat p(a)-\widehat p(a_0))-(p(a)-p(a_0))\right|.
\end{equation}
State-wise offset correction cancels in this difference away from clipping.
We also report height effects, Spearman correlation with its defined fraction,
and local constrained-selection regret. Local choice maximizes
$\mathrm{clip}(\widehat h_5/.10,0,1)$ among candidates whose predicted
maximum peak is at most 8 N, without the controller's force/action costs.
Exact ties prefer the center then ascending index; if none is feasible,
it chooses minimum predicted maximum force. Executed utility is
$\mathrm{clip}(h_5/.10,0,1)$ if the branch itself respects 8 N, otherwise
zero; prebranch violations are excluded. Constant predictions have undefined
correlation. Regret is interpreted with oracle utility, since zero regret
can be trivial when all candidates have zero utility. Bootstrap resampling
retains all prefixes, branches, horizons, and model seeds within an environment.
All prefix replay differences were zero. Figure~\ref{fig:effects} separates absolute accuracy from action sensitivity. Raw VT and current-normal persistence have nearly identical absolute peak MAE (0.371 versus 0.371 N). Yet VT reduces force action-effect MAE from 0.413 to 0.338 N: paired difference \mbox{$-0.074\,[-0.087,-0.062]$} N. Height-effect MAE also decreases from 0.864 to 0.396 mm. Raw vision force-effect MAE is 1.408 N. Anchoring changes VT absolute/effect peak MAE to 0.364/0.336 N, a small numerical change rather than new learned dynamics.

Raw VT peak-rank correlation is 0.894, defined on 66.7\% of prefix states; persistence rankings are undefined. Mean oracle local utility is 0.358. Persistence regret is 0.01080, while raw vision and VT regrets are below $10^{-5}$. All 324 executed branches respect 8 N; oracle utility averaged by prefix is 0, 0.073, and 1. Zero early utility and late clipping limit regret discrimination. The grid contains 288 distinct state--action branches after clipping duplicates. This diagnostic supports local action-effect information, but barely distinguishes learned selectors and supplies no evidence of superior force protection. Five-step choice is also distinct from sustained, full-episode lifting.

\subsection{Public tactile-error perturbation}
The public TacBench GelSight Mini/ATI Nano17 records~\cite{sparsh} pair
optical contact images with forces, not this model's RGB/action trajectories.
A separate 361,091-parameter CNN uses 453 training and 75 validation
trajectories. We fix its first seed, 17, and reuse ordered held-out ID
prediction-minus-label normal-force residuals (116 trajectories; 9,003 frames).
Of these, 47 trajectories admit 93-frame blocks, with 951 possible starts.
We sample an eligible trajectory uniformly, then a start uniformly within
it, without wrapping or rescaling. Independent finger blocks are shared
across all methods on each of the same 30 main ID environments.

At observations 42--134, channels 0 and 3 become
$\max(0,F_n+e_n)$; tangential channels are unchanged. All tactile consumers,
including encoders, feedback, guards, and anchoring, receive this stream.
Errors are not contact-gated: adding contact-derived residuals at zero true
force can create synthetic false contacts. Physics is not directly perturbed,
but altered actions can change its trajectory; outcome metrics use true forces
and heights. Eleven fixed configurations give 330 executions. Paired
stress-minus-clean intervals use 2,000 environment-bootstrap draws (seed
20401010), conditional on the fitted policies and residual bank, with
training-seed variation reported separately.

The source bank's normal labels span approximately 0.073--1.888 N, below
the 8 N simulator budget. One source residual per 20 Hz observation is
an artificial index alignment because source timestamps are unavailable.
Independent blocks from one sensor do not identify two-finger error
correlation. These regression residuals include prediction and reference
measurement error; they are not isolated sensor noise.
WM-RL VT joint success changes by \mbox{$-1.1\,[-3.3,+0.0]$} pp; IQL VT changes by \mbox{$+1.1\,[-2.2,+5.6]$} pp. Model-assisted feedback, force feedback, and the reactive base retain their clean joint-success and force-violation rates. The narrow source-force range and this fixed perturbation bank make unchanged outcomes weak evidence of general robustness; zero empirical bootstrap width does not prove zero population uncertainty. 
This tests sensitivity to one empirical error sequence distribution;
it does not demonstrate optical-sensor transfer or physical-robot control.

\section{Discussion and Limitations}
The evidence supports conditional value for the tested compact model. Its action-effect predictions outperform a zero-effect persistence reference, and its local selector improves whole-episode ID and geometry-shift feedback outcomes through fewer lowering-phase violations. Those findings coexist with no demonstrated pooled joint-success advantage and substantial lifting losses under physical-parameter shifts. The guard ablations and error perturbations add useful controls, but sparse guard activations and small perturbations do not establish protection under challenging disturbances.

Imagined optimization increases height completion over BC without a demonstrated joint-success gain, and reactive IQL performs better in the pooled primary outcome. A dense height reward also differs from the strict ten-observation lift and whole-episode force event. These comparisons do not identify a single failure mechanism or show that all world-model RL methods are inferior; they establish the limits of this frozen-model pipeline and why an independent offline learner is a necessary reference.

The common privileged approach, two controlled action dimensions,
five rigid-box geometries, and short-horizon local branches restrict
generalization. Analytic friction/aperture screening removes a simple
capacity confound but does not establish dynamic feasibility. Three
training seeds and 30 environments per domain leave uncertainty in
small effects and heterogeneous failures. Absence of a significant
difference is not evidence of practical equivalence.

The world-model pipeline uses predictive auxiliary supervision and a
different optimization budget from reactive IQL; its comparison is
not a one-factor intervention. Model removal holds the feedback base
fixed, but there is no affine action-conditioned predictor, tactile-only
model, action-shuffled planner, or capacity sweep. We therefore avoid
claims that deep multimodal dynamics are universally necessary or
computationally superior. Neither the measured guard, predicted
feasibility, nor recording-based calibration guarantees closed-loop
safety. Rigid contact forces also do not measure deformation or damage.

\section{Conclusion}
Matched control executions and independently replayed action branches separate sensing accuracy, local dynamics, and force-budgeted lifting. The compact visuotactile model supplies useful local action-effect information and improves a feedback controller in ID and geometry-shift conditions. Imagined policy learning and nominal recording-based calibration do not establish a general constrained-control advantage. Reporting persistence, independent IQL, paired joint outcomes, and physical-parameter shifts makes that distinction visible.

\section*{Reproducibility and Tool Use}
The simulator uses robosuite 1.5.1 and MuJoCo 3.3.7 with fixed seeds and saved
episode parameters. Source, training logs, per-environment outcomes,
protocol amendments, and analysis scripts retain the original exploratory
study separately. TD-MPC2 components retain their MIT license; the
public tactile images are not redistributed. Derived residual blocks retain
their source CC BY-NC 4.0 terms separately from software. Training used an RTX 5090 Laptop GPU (24 GB), PyTorch 2.14.0+cu130. Observed per-run wall times were 147--212 s for world models including final evaluation, 53--57 s for BC plus imagined RL, and 663--832 s for reactive BC plus IQL. Concurrent execution prevents an isolated speed comparison. Branch forecasts used CPU inference after an initialization GPU-memory failure; training and primary learned-policy inference used GPU.

OpenAI Codex assisted with drafting and revising text throughout this
manuscript, developing the experiment and analysis code underlying
Sections III--V, preparing the figures and tables, and orchestrating
the reported experiments.


\begin{thebibliography}{12}
\bibitem{tactilempc}
S.~Tian, F.~Ebert, D.~Jayaraman, M.~Mudigonda, C.~Finn, R.~Calandra,
and S.~Levine, ``Manipulation by feel: Touch-based control with deep
predictive models,'' in \emph{Proc. IEEE Int. Conf. Robot. Autom.}, 2019.
\url{https://arxiv.org/abs/1903.04128}.
\bibitem{vtwm}
C.~Higuera, S.~Arnaud, B.~Boots, M.~Mukadam, F.~R.~Hogan, and F.~Meier,
``Visuo-tactile world models,'' arXiv:2602.06001v1, 2026.
\bibitem{feelworld}
W.~Ma, C.~Zhang, C.~Xue, Y.~Cai, G.~Yao, S.~Cui, and S.~Wang,
``FeelWorld: Visuo-tactile world model for hierarchical contact prediction
and planning,'' arXiv:2607.24267v1, 2026.
\bibitem{omnivta}
Y.~Zheng \emph{et al.}, ``OmniVTA: Visuo-tactile world modeling for
contact-rich robotic manipulation,'' arXiv:2603.19201v3, 2026.
\bibitem{tdmpc2}
N.~Hansen, H.~Su, and X.~Wang, ``TD-MPC2: Scalable, robust world models
for continuous control,'' in \emph{Proc. Int. Conf. Learn. Representations},
2024. \url{https://www.tdmpc2.com/}.
\bibitem{dreamer}
D.~Hafner, J.~Pasukonis, J.~Ba, and T.~Lillicrap, ``Mastering diverse
control tasks through world models,'' \emph{Nature}, vol.~640,
pp.~647--653, 2025, doi: 10.1038/s41586-025-08744-2.
\bibitem{iql}
I.~Kostrikov, A.~Nair, and S.~Levine, ``Offline reinforcement learning
with implicit Q-learning,'' arXiv:2110.06169, 2021.
\bibitem{valueequivalence}
C.~Grimm, A.~Barreto, S.~Singh, and D.~Silver, ``The value equivalence
principle for model-based reinforcement learning,'' in \emph{Advances in
Neural Information Processing Systems}, 2020.
\bibitem{conformal}
A.~N.~Angelopoulos and S.~Bates, ``Conformal prediction: A gentle
introduction,'' \emph{Found. Trends Mach. Learn.}, vol.~16, no.~4,
pp.~494--591, 2023, doi: 10.1561/2200000101.
\bibitem{sparsh}
C.~Higuera \emph{et al.}, ``Sparsh: Self-supervised touch representations
for vision-based tactile sensing,'' in \emph{Proc. Conf. Robot Learning
(CoRL 2024)}, Proc. Mach. Learn. Res., vol.~270, pp.~885--915, 2025.
Force data: \url{https://huggingface.co/datasets/facebook/gelsight-force-estimation}.
\bibitem{robosuite}
Y.~Zhu \emph{et al.}, ``robosuite: A modular simulation framework and
benchmark for robot learning,'' arXiv:2009.12293v3, 2025. Original preprint, 2020.
\bibitem{mujoco}
E.~Todorov, T.~Erez, and Y.~Tassa, ``MuJoCo: A physics engine for
model-based control,'' in \emph{Proc. IEEE/RSJ Int. Conf. Intell. Robots Syst.},
2012, doi: 10.1109/IROS.2012.6386109.
\end{thebibliography}
\end{document}